\documentclass[runningheads]{llncs}
\usepackage[T1]{fontenc}
\usepackage{graphicx}
\usepackage{bm}
\usepackage{amsmath} 
\usepackage{amsfonts} 
\usepackage{makecell}
\usepackage{epsfig}
\usepackage{graphicx,epstopdf}
\usepackage[misc]{ifsym}
\usepackage[hidelinks,colorlinks=true,linkcolor=blue,citecolor=blue,urlcolor=blue]{hyperref}

\begin{document}
\title{Biomechanics-informed Non-rigid Medical Image Registration and its Inverse Material Property Estimation with Linear and Nonlinear Elasticity
}
\author{Zhe Min  \inst{1,2}\textsuperscript{(\Letter)} \and
  Zachary M. C. Baum \inst{2} \and
  Shaheer U. Saeed \inst{2}
  \and
  Mark Emberton \inst{3}
\and \\
Dean C. Barratt \inst{2}\and 
 Zeike A. Taylor  \inst{4}
 \and
 Yipeng Hu \inst{2}
 }
\institute{
School of Control Science and Engineering, Shandong University, Jinan, China
\email{minzhe@sdu.edu.cn;z.min@ucl.ac.uk}
\and
Centre for Medical Image Computing and Wellcome/EPSRC Centre for
Interventional \& Surgical Sciences,  University College London, London, UK
\and 
Division of Surgery \& Interventional Science, University College London, London,
UK
\and
CISTIB Centre for Computational Imaging and Simulation Technologies in
Biomedicine, Institute of Medical and Biological Engineering, University of Leeds,
Leeds, UK
}

\authorrunning{Z. Min et al.}
\maketitle

\begin{abstract}
This paper investigates both biomechanical-constrained non-rigid medical image registrations and accurate identifications of material properties for soft tissues, using physics-informed neural networks (PINNs). The nonlinear elasticity theory is leveraged to formally establish the 
partial differential equations (PDEs) representing physics laws of biomechanical constraints to be satisfied, with which registration and identification tasks are treated as forward (i.e., data-driven solutions of PDEs) and inverse (i.e., 
parameter estimation) problems under PINNs respectively. While the forward problem has direct clinical applications in guiding targeted biopsy and treatment, the solution to the inverse problem may open new research directions in quantifying disease-indicative mechanical properties of \textit{in vivo} tissues. Two net configurations (i.e., Cfg1 and Cfg2) have also been compared for both linear and nonlinear physics models, according to whether backbones are shared between branches or not. Two sets of experiments have been conducted, using pairs of undeformed and deformed MR images from clinical cases of prostate cancer biopsy. In the first experiment, against the finite-element-computed ground-truth, the root mean squared error (rmse) of registration for surface points 
was reduced from $1.83\pm0.51$ mm without PINNs to $1.43\pm0.70$ mm (Cfg1, $p=0.024$) and 
$1.23\pm0.69$ mm (Cfg2, $p<0.001$) with linear elasticity, and to $1.45\pm0.84$ mm (Cfg1, $p=0.004$) and $1.24\pm0.69$ mm (Cfg2, $p<0.001$) with nonlinear elasticity, while average differences between linear and nonlinear models were not found statistically significant (e.g., $p=0.972$ between two Cfg1s) but their respective benefits may depend on specific patients.
In the second experiment, the nonlinear model exhibited evident advantages over the linear counterpart ($p=0.002$) in predicting ratios of tissue stiffness (i.e., Young's modulus) between two subregions (i.e., peripheral and transition zones) of the prostate, with the mean average percentage error (mAPE) values being $14.20\%\pm14.12\%$ and $76.28\%\pm30.97\%$, respectively. The codes are available at \url{https://github.com/ZheMin-1992/Registration_PINNs}. 
\end{abstract}
\section{Introduction}
Biomechanical modelling plays an important role in regularising medical image registration \cite{fu2021biomechanically,xu2022double,xu2021f3rnet} that further enables surgical guidance for different organs (e.g., prostate \cite{fu2021biomechanically,hu2018weakly,van2015biomechanical,zeng2020label}, liver \cite{pfeiffer2020non}, brain \cite{luo2020registration,luo2022dataset} and heart \cite{QIN2023102682}), for example, to constrain predicted spatial transformations to be physically plausible, under either iterative optimisation \cite{van2015biomechanical} or neural-network-training \cite{hu2018weakly,zeng2020label} schemes. Biomechanical constraints could vary from simple linear \cite{linearandnonlinear_elasticity} to complex nonlinear models \cite{fu2021biomechanically,viscoelasticity_model_reference3}, while they require values of material properties if applied to the registration problem. This paper investigates both aspects by leveraging the capabilities of physics-informed neural networks (PINNs) to seek data-driven solutions (i.e., forward problem) and enable data-driven discovery (i.e., inverse problem) of partial differential equations (PDEs)  \cite{pinns_review} respectively, in the non-rigid medical image registration and material property estimation. \\
\indent Linear elasticity models assuming a linear relationship between stress and strain, are only effective for modelling small deformations under low-stress conditions \cite{linearandnonlinear_elasticity}. Nonlinear elasticity models built on more complex constitutive models and strain energy functions, are better suited for capturing large deformations and nonlinear material behavior \cite{linearandnonlinear_elasticity}. For example, the anisotropic viscoelasticity constitutive models \cite{viscoelasticity_model_reference2,viscoelasticity_model_reference1,viscoelasticity_model_reference3} were utilised for simulating soft tissues' deformations, led to more realistic organ (e.g., liver) geometries than linear models with desired properties such as stress dissipation \cite{taylor2009modelling}. The nonlinear finite-element-analysis was developed for modelling-fidelity surgical simulations \cite{taylor2008high}.  However, the choices of hyperviscoelastic, hyperelastic, and linear elastic constitutive models were demonstrated to be not important in estimating brain shifts for an image-guided neurosurgery procedure \cite{wittek2009unimportance}. \textit{It is unclear whether a more complex nonlinear model (i.e., both geometrical and constitutive models) could make a difference in the non-rigid point set registration problem.}\\
\indent 
Material property estimation aims to quantify mechanical properties of materials (e.g., soft tissues), which can be used to develop customized surgical plans that could improve surgical outcomes \cite{hu2012mr,hu2011modelling}. Perhaps more interestingly, it may assist disease detection and localisation, as studies frequently found \textit{in vitro} tumour has distinct mechanical properties compared to the healthy tissue \cite{ji2019stiffness}.
The identifications of bulk and shear modulus, under both linear elasticity and hyperelasticity models, were explored using specimens with simple one-or-two layers structures through experimental tests (e.g., uniaxial tension or compression) \cite{hartmann2018identifiability}. The authors revealed multiple findings about under what conditions parameters are identifiable, for example, to compute material parameters of a single-layered specimen, a uniaxial tensile test either together with lateral strain measurement, with torsion or with a shear experiment is needed \cite{hartmann2018identifiability}. We note that circumstances are much more complicated in both soft-tissue deformation modelling where exact values of stress and strain are unknown, and image registration where displacement vectors of voxels need to be estimated. \textit{It is thus unknown whether material properties are identifiable or not, within the challenging scenario of image registration where boundary conditions also need to be
estimated.}
\\
\indent 
To answer the first question of whether nonlinear elasticity models are better than linear counterparts in registration, in a similar fashion with a recent work \cite{min2023non} where linear elasticity was adopted, this study incorporates nonlinear biomechanical constraints in forms of PDEs into a PINN for registration. The findings first confirmed the validity of adopting PINNs to impose biomechanical constraints for predicted transformations and to reduce registration error, and also indicated that there existed no statistically significant difference between average performances using linear and nonlinear models while choices may depend on specific patients. 
To answer the second question of whether material properties are identifiable along with registration, PINNs were utilised to formulate the joint optimisation framework of the registration problem and the material property estimation problem. The results demonstrated that ratios of soft tissues' stiffness (i.e., Young's modulus) between two distinct compartments (i.e.,the peripheral zone and transition zone) of the prostate gland could be successfully recovered, while nonlinear models exhibited evident advantages over their linear counterparts in this problem. 
 \\
\indent Our contributions are summarised as follows. \textbf{1)} We developed a learning-based biomechanical-constrained non-rigid registration algorithm using PINNs, where linear elasticity is generalised to the nonlinear version. \textbf{2)} We demonstrated extensively that nonlinear elasticity shows no statistical significance against linear models in computing point-wise displacement vectors but their respective benefits may depend on specific patients, with finite-element (FE) computed ground-truth. \textbf{3)} We formulated and solved the inverse parameter estimation problem, under the joint optimisation scheme of registration and parameter identification using PINNs, whose solutions can be accurately found by locating saddle points of the optimisation. 
\section{Methods}
\label{methods}
\begin{figure}[t]
\includegraphics[width=\textwidth]
{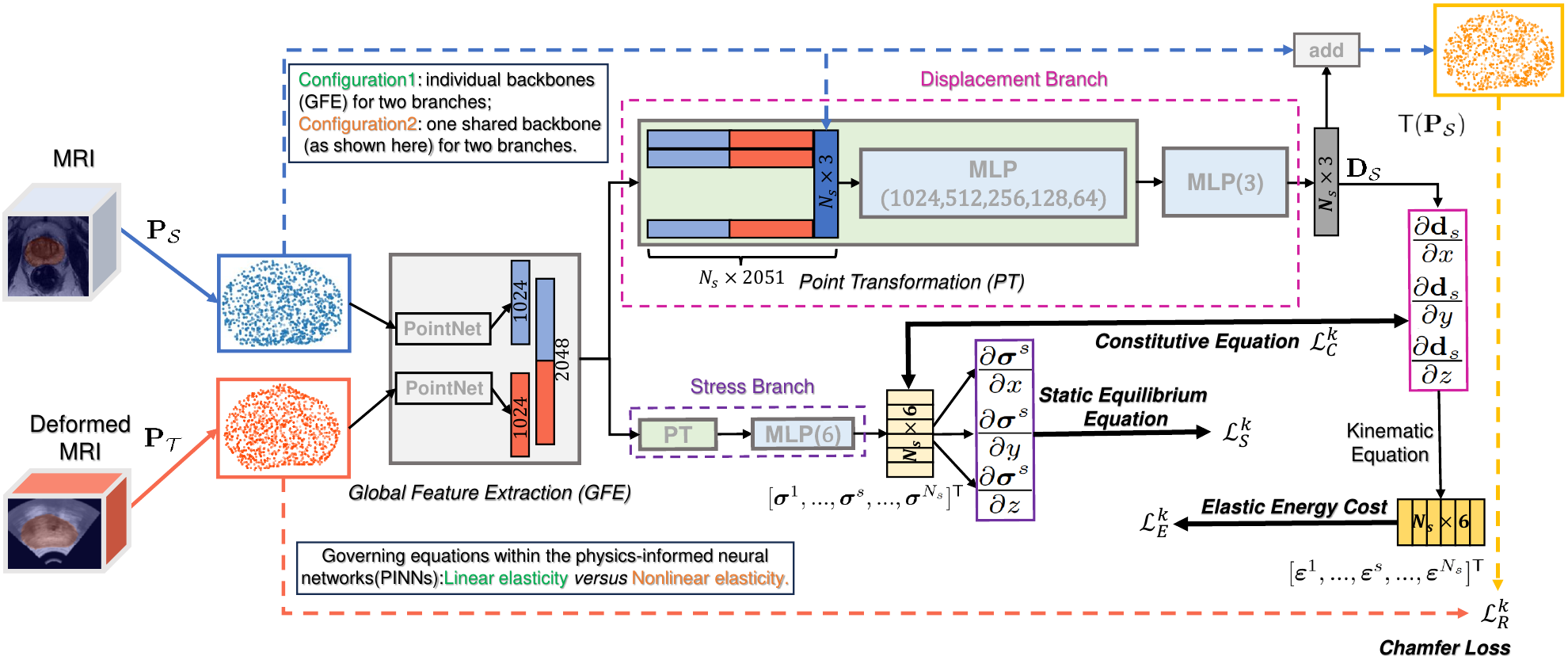}
 \caption{ The schematic of the proposed physics-informed neural networks.
 }
\label{Illustration_method}
\end{figure}

\indent  The non-rigid 3D point set registration problem in fields of medical imaging, is to warp the source point set
$\mathbf{P}_\mathcal{S}\in\mathbb{R}^{N_s\times3}$ with $\mathbf{p}_s\in\mathbb{R}^3$
to the target point set $\mathbf{P}_\mathcal{T}\in\mathbb{R}^{N_t\times3}$ with $\mathbf{p}_t\in\mathbb{R}^3$, so as to accurately map important anatomical structures in two spaces. The warped source point set is $\mathsf{T}(\mathbf{P}_\mathcal{S}) = \mathbf{P}_\mathcal{S} + \mathbf{D}_\mathcal{S} $, where the displacement vectors are $\mathbf{D}_\mathcal{S}\in\mathbb{R}^{N_s\times3}$ with $\mathbf{d}_s\in\mathbb{R}^3$.\\
 \indent The idea of imposing biomechanical constraints in the registration problem with physics-informed neural networks (PINNs), like that in \cite{min2023non}, is to predict point-wise displacement vectors and biomechanical-related values (e.g., stress or strain), of which the underlying physics laws (i.e., governing equations) usually represented by partial differential equations (PDEs) should be satisfied. 
 As shown in Fig. \ref{Illustration_method} and introduced in details in Sect. \ref{the physical laws to be satisfied}, there are three main governing equations and one energy function in modelling deformation of soft tissues, from simple linear relationships \cite{min2023non}, to the elaborate nonlinear cases (i.e., compressible Neo-Hookean model \cite{ogden2013non}) in this study. Let 
 $\mathcal{D}_k$ be the data set of $k$-th patient, containing $\mathbf{P}_\mathcal{S}$ and $\mathbf{P}_\mathcal{T}$, 
 Fig. \ref{Illustration_method} shows the algorithm schematic where the displacement-predicting branch is $g_{\theta_g}(\mathcal{D}_k  
)$ and the stress-predicting branch is $h_{\theta_h}(\mathcal{D}_k)$ with learnable parameters $\theta_g$ and $\theta_h$, which together constitutes $e_{\theta}(\mathcal{D}_k)$ with $\theta$. Two net configurations have been compared, where individual and shared (as depicted in Fig. \ref{Illustration_method}) backbones (i.e., global feature extraction module based on PointNet \cite{qi2017pointnet}) are utilised to extract features for two branches respectively. For clarity, the four models with two net configurations and two physics models are denoted as \textsf{Linear Cfg1}, \textsf{Linear Cfg2}, \textsf{Nonlinear Cfg1} and \textsf{Nonlinear Cfg2}.
\subsection{The Forward Problem of Non-Rigid Point Set Registration using PINNs }
\label{non-rigid point set registration using PINNs}
\indent 
The registration task is formulated as the \textsf{forward} problem (i.e., 
the data-driven solutions of PDEs) which estimates the unknown function $g_{\theta_g}(\mathcal{D}_k  
)$ within the PINNs framework where nonlinear PDEs are parameterised by lame parameters $\lambda_s\in \mathbb{R}$ and $\mu_s\in \mathbb{R}$ at the point $\mathbf{p}_s$. Here, boundary conditions are `actual' displacement vectors of source points which are unknown and thus their estimation errors are approximated with the Chamfer loss. The Chamfer loss $\mathcal{L}_{R}^k(\theta_g;\mathcal{D}_k)$ first seeks the nearest point in the other point set (e.g., $\mathsf{T}(\mathbf{P}_\mathcal{S})$) to each point in one point set(e.g., $\mathbf{P}_\mathcal{T}$) and computes the $\text{L}2$ distance, repeats the above operation for the other point set $\mathsf{T}(\mathbf{P}_\mathcal{S})$ and returns the sum of average distances.  The Chamfer loss can be either applied to all points $\mathbf{P}_\mathcal{S}$ 
and $\mathbf{P}_\mathcal{T}$ 
or surface points only
$\mathbf{P}_\mathcal{S}^{\text{surface}}\in\mathbb{R}^{N_s^{\text{surface}}\times3}$ and
$\mathbf{P}_\mathcal{T}^{\text{surface}}\in\mathbb{R}^{N_t^{\text{surface}}\times3}$ (i.e., `strict' boundary conditions where surface points are points on the boundary). By regarding all points as `collocation' points following PINNs notation conventions \cite{raissi2019physics}, as shown in Fig. \ref{Illustration_method}, nonlinear PDEs include the static equilibrium deviation loss $\mathcal{L}_{S}^k(\theta_h;\mathcal{D}_k)=\sum_{s=1}^{N_s} f_{\text{static}}(\frac{\partial \bm{\sigma}^s }{\partial x},\frac{\partial \bm{\sigma}^s }{\partial y}, \frac{\partial \bm{\sigma}^s }{\partial z})$ that regularizes spatial derivatives of stress $\bm{\sigma}^s$, the constitutive deviation loss $\mathcal{L}_{C}^k(\theta;\mathcal{D}_k)= \sum_{s=1}^{N_s} 
       f_{\text{const}}(\bm{\sigma}^s, \frac{\partial \mathbf{d}_s} {\partial \mathbf{p}_s},  \lambda_s,\mu_s)$ that relates $\bm{\sigma}^s$ and spatial derivatives of displacements $\frac{\partial \mathbf{d}_s} {\partial \mathbf{p}_s}$ with  $\lambda_s$ and $\mu_s$, and the elastic energy loss $ \mathcal{L}_{E}^k(\theta_h;\mathcal{D}_k) = \sum_{s=1}^{N_s} 
   f_{\text{energy}}(\mathbf{\bm{\varepsilon}}^s, \lambda_s,\mu_s)$ that relies on strains and lame parameters. Note that $f_{\text{static}}(\star)$, $f_{\text{const}}(\star)$ and $f_{\text{energy}}(\star)$ will be defined in Sect. \ref{the physical laws to be satisfied}. The training loss $\mathcal{L}^k(\theta;\mathcal{D}_k)$ in the forward problem for the $k$-th subject is given by a  ($w\in\mathbb{R}^{+}$)-weighted sum of these terms, 
\begin{equation}
\label{the overall loss function}
\mathcal{L}^k(\theta;\mathcal{D}_k) =
    w\mathcal{L}_{R}^k(\theta_g;\mathcal{D}_k)
    + \mathcal{L}_{S}^k(\theta_h,\mathcal{D}_k)
    + \mathcal{L}_{C}^k(\theta;\mathcal{D}_k)
    +\mathcal{L}_{E}^k(\theta_h;\mathcal{D}_k),
\end{equation}
where $\mathcal{D}_k$ is considered to include lam\'{e} parameters $\lambda_s$ and $\mu_s$.
\subsection{The Inverse Problem of Identifying Material Properties of Soft Tissues using PINNs}
\label{the inverse problem section}
In Sect. \ref{non-rigid point set registration using PINNs}, we treat the registration as the \textsf{forward} problem using PINNs, which necessitates knowing material properties (e.g., Young's modulus).  In contrast, the parameter (e.g., material properties) estimation is regarded as the \textsf{inverse} problem (i.e., 
the data-driven discovery of PDEs) under a similar PINNs framework. 
Consider two distinct compartments of the prostate gland, i.e., the peripheral zone (PZ) and transition zone (TZ), which exhibit different stiffness magnitudes \cite{hu2011modelling,hu2012mr}. Let $\mathcal{D}_k^{\text{PZ}}$ and $\mathcal{D}_k^{\text{TZ}}$ denote points in  PZ and TZ, $E_k^{\text{PZ}}\in \mathbb{R}$ and $E_k^{\text{TZ}}\in \mathbb{R}$ be their respective Young's modulus values.  
The particular example of the inverse problem here, assuming that $E_k^{\text{TZ}}$ is known, is to estimate the ratio $\beta_k\in \mathbb{R}$ of $E_k^{\text{PZ}}$ to $E_k^{\text{TZ}}$, which plays an important role in biomechanics-constrained non-rigid point set registration as verified in \cite{hu2011modelling}.
 To this end, one learnable weight that functions as $\beta_k$ is added into the network $e_{\theta}(\mathcal{D}_k)$. The loss function for the inverse problem, under the joint learning scheme of point set registration and parameter estimation, is
\begin{equation}
   \label{the overall loss function also considering the parameter estimation as well}
\mathcal{L}^k(\theta, \beta_k;\mathcal{D}_k) =
    w\mathcal{L}_{R}^k(\theta_g;\mathcal{D}_k)
    + \mathcal{L}_{S}^k(\theta_h,\mathcal{D}_k)
    + \mathcal{L}_{C}^k(\theta,\beta_k;\mathcal{D}_k)
    +\mathcal{L}_{E}^k(\theta_h,\beta_k;\mathcal{D}_k),
\end{equation}
where $\mathcal{L}_{C}^k(\theta,\beta_k;\mathcal{D}_k)$ and 
$\mathcal{L}_{E}^k(\theta,\beta_k;\mathcal{D}_k)$ can be expanded as
$\mathcal{L}_{C/E}^k(\theta,\beta_k;\mathcal{D}_k) \equiv 
    \mathcal{L}_{C/E}^k(\theta;\mathcal{D}_k^{\text{TZ}}, E_k^{\text{TZ}})  + 
     \mathcal{L}_{C/E}^k(\theta;\mathcal{D}_k^{\text{PZ}}, \beta_k E_k^{\text{TZ}}),
$ where lame parameters $\lambda_s$ and $\mu_s$ at $\mathbf{p}_s$ are computed using $E_k^{\text{TZ}}$ and $\beta_k E_k^{\text{TZ}}$ depending on the sub-regions (i.e., $\mathcal{D}_k^{\text{PZ}}$ or $\mathcal{D}_k^{\text{TZ}}$) in which $\mathbf{p}_s$ falls.  \\
\indent The optimisation problem minimising $\mathcal{L}^k(\theta, \beta_k;\mathcal{D}_k)$ in Eq. (\ref{the overall loss function also considering the parameter estimation as well}) is likely to be ill-posed, in that it contains unknowns in both registration-related and physics-laws-related loss terms (i.e., PDEs). Instead of seeking local minimums w.r.t. material parameters which is likely to be
unidentifiable and produces naive solutions, we investigate saddle points that are much more interesting and represent practically meaningful solutions as will be demonstrated in Sect. \ref{Experiments and Results}. 
\subsection{Governing Equations for Deforming Soft Tissues Considering Nonlinear Elasticity}
\label{the physical laws to be satisfied}
 \indent In this section, we describe in details PDEs representing governing  equations in both the forward problem in Sect. \ref{non-rigid point set registration using PINNs} and the inverse problem in Sect. \ref{the inverse problem section}. \\
\noindent \textbf{Nonlinear Strain-displacement Equations} The nonlinear strain-displacement equation at a source point $\mathbf{p}_s$ is 
\begin{equation}
\label{nonlinear kinematic equation}
\mathbf{\bm{\varepsilon}}^s = \frac{1}{2}( \nabla \mathbf{d}_s + \nabla \mathbf{d}_s^{\mathsf{T}}
+ \nabla \mathbf{d}_s^{\mathsf{T}}\nabla \mathbf{d}_s
),
\end{equation}
where $\bm{\varepsilon}^s$ is the \textit{Green-Lagrangian strain} tensor at $\mathbf{p}_s$, $\nabla \mathbf{d}_s$ is the 
displacement gradient w.r.t. spatial coordinates $x,y,z$ of $\mathbf{p}_s$. As shown in Fig. \ref{Illustration_method}, Eq. (\ref{nonlinear kinematic equation}) is utilised to derive point-wise strain $\bm{\varepsilon}^s$ from the predicted displacement vector $\mathbf{d}_s$. \\
\noindent \textbf{Static Equilibrium Equations} In nonlinear elasticity, $\bm{\sigma}^s$ predicted by $h_{\theta_h}(\mathcal{D}_k)$ at $\mathbf{p}_s$ would be a \textit{2nd Piola-Kirchhoff stress} tensor, satisfy the following equilibrium equation $\sigma_{ji,j}^s+ F_{i}=0$ where 
$(\cdot)_{,j}$ is a shorthand for $\frac{\partial{(\cdot)}}{
\partial(\mathbf{p}_s)_j
}$
 , $F_i\in\mathbb{R}$ is the body force that is assumed to be zero at the static equilibrium, $i$ and $j$ denote three spatial directions. The PDEs that compose $\mathcal{L}_{S}^k(\theta_h;\mathcal{D}_k)$ in Eqs. (\ref{the overall loss function}) and (\ref{the overall loss function also considering the parameter estimation as well}) are $f_{\text{static}}(\frac{\partial \bm{\sigma}^s }{\partial x},\frac{\partial \bm{\sigma}^s }{\partial y}, \frac{\partial \bm{\sigma}^s }{\partial z}) =
||    \frac{\partial \sigma_{xx}^s}{\partial x}
    +\frac{\partial \sigma_{yx}^s}{\partial y}
    +\frac{\partial \sigma_{zx}^s}{\partial z}||_2^2 +||
      \frac{\partial \sigma_{xy}^s}{\partial x}
    +\frac{\partial \sigma_{yy}^s}{\partial y}
    +\frac{\partial \sigma_{zy}^s}{\partial z}||_2^2 
    + ||\frac{\partial \sigma_{xz}^s}{\partial x}
    +\frac{\partial \sigma_{yz}^s}{\partial y}
    +\frac{\partial \sigma_{zz}^s}{\partial z}||_2^2$. \\
\noindent \textbf{Nonlinear Constitutive Equations}
The stress and displacement gradients are further constrained by the constitutive equation 
as $\bm{\sigma}^s =   \mu_s  J_s^{-1} (\mathbf{F}_s\mathbf{F}_s^{\mathsf{T}} -\mathbf{I}_{3\times3})
+ \lambda_s (J_s-1)\mathbf{I}_{3\times3}$, where $\mathbf{F}_s= \mathbf{I}_{3\times3} + \frac{\partial \mathbf{d}_s} {\partial \mathbf{p}_s}\in \mathbb{R}^{3\times 3}$ is the deformation gradient at $\mathbf{p}_s$, $J_s = \text{det}(\mathbf{F}_s) \in \mathbb{R}$ is the determinant of $\mathbf{F}_s$, $\lambda_s \in \mathbb{R}$ and $\mu_s \in \mathbb{R}$ 
are Lame parameters at $\mathbf{p}_s$ which are computed with Young's Modulus $E_s$ and Possion's ratio $v_s$ using $\lambda_s = \frac{E_s\nu_s}{(1-2\nu_s) (1+\nu_s)}$ and  $\mu_s =\frac{E_s}{2(1+\nu_s)}$.
The PDEs that comprise $\mathcal{L}_{C}^k(\theta;\mathcal{D}_k)$ in Eqs. (\ref{the overall loss function}) and (\ref{the overall loss function also considering the parameter estimation as well}) are 
$f_{\text{const}}(\bm{\sigma}^s,\frac{\partial \mathbf{d}_s} {\partial \mathbf{p}_s}, \lambda_s,\mu_s)=\sum_{i\in\{1,2,3\}}||\sigma_{ii}^{s} - \mu_s J_s^{-1} ((\mathbf{F}_s\mathbf{F}_s^{\mathsf{T}})_{ii}-1)+\lambda_s(J_s-1)||_2^2+\sum_{\langle i,j\rangle \in \{ 
  \langle 1,2\rangle, \langle 1,3\rangle, \langle 2,3 \rangle       \}}
     ||\sigma_{ij}^s -\mu_s J_s^{-1}(\mathbf{F}_s\mathbf{F}_s^{\mathsf{T}})_{ij}||_2^2$, where to maintain uniformity of notations here $\sigma_{11}^{s}=\sigma_{xx}^{s}$ (similar for index pairs $\langle2,2\rangle$ and $\langle3,3\rangle$) and $\sigma_{12}^{s}=\sigma_{xy}^{s}$ (similar for index pairs $\langle1,3\rangle$ and $\langle2,3\rangle$) hold. \\
\noindent \textbf{Nonlinear Elastic Energy Density Function} The elastic energy function $f_{\text{energy}}(\mathbf{\bm{\varepsilon}}^s,\lambda_s,\mu_s)$ that forms  $\mathcal{L}_{E}^k(\theta_h;\mathcal{D}_k)$ in Eqs. (\ref{the overall loss function}) and (\ref{the overall loss function also considering the parameter estimation as well}) is $f_{\text{energy}}(\mathbf{\bm{\varepsilon}}^s, \lambda_s,\mu_s) = \frac{\mu_s}{2} \Big( \text{tr}(\mathbf{F}_s\mathbf{F}_s^{\mathsf{T}})
    -3-2\ln J_s \Big) + \frac{\lambda_s}{2}\big( J_s-1\big)^2
    \big)$ with $\text{tr}(\mathbf{F}_s\mathbf{F}_s^{\mathsf{T}})$ represented by $\mathbf{\bm{\varepsilon}}^s$.
\section{Experiments and Results}
\label{Experiments and Results}
\begin{figure}[t!]
 \begin{tabular}{c}          
 \includegraphics[width=1\linewidth]
{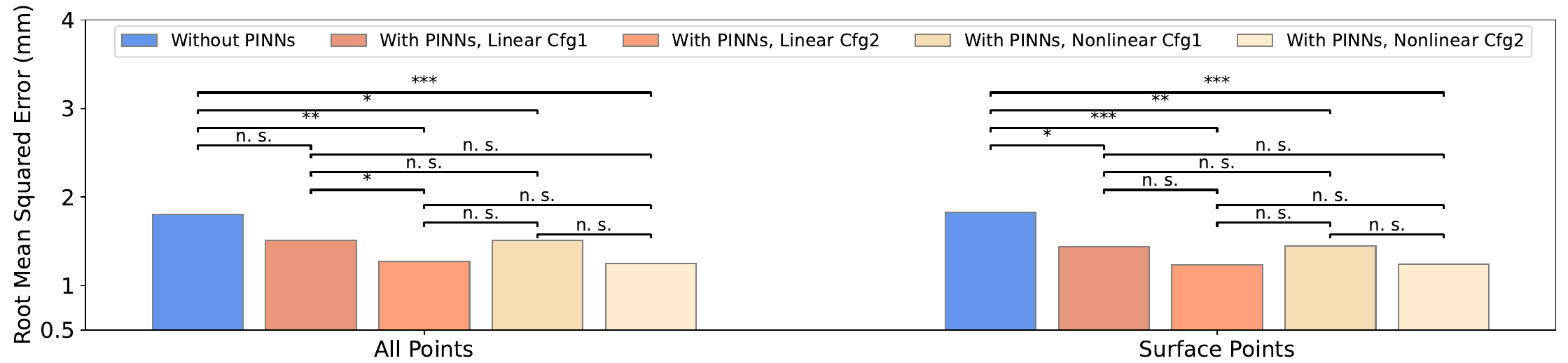}
\end{tabular}
\caption{The root-mean-squared-error (rmse) of registration using different algorithms for surface and all points on the left and right subplots, respectively. 
n.s.: not significant, $\star:p<0.05$, $\star\star:p<0.01$, $\star\star\star:p<0.001$.
} 
\label{quantitative results of linear and nonlinear}
\end{figure}

\begin{figure}[t]
 \begin{tabular}{c}          
\includegraphics[width=1\linewidth]{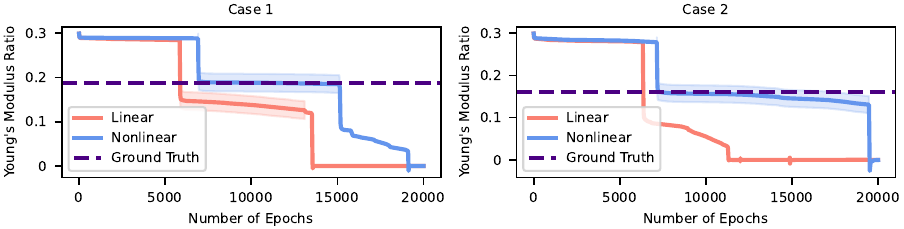}
\end{tabular}
\caption{The optimisation process of the inverse problem estimating the young's modulus ratio between two sub-regions of prostate, are found by locating saddle points.}
\label{Inverse estimation optimisation process}
\end{figure}

\textbf{Data Sets and Evaluation Metrics} 
The dataset contains $N_k=22$ pairs of point sets generated over MRI-derived prostate meshes \cite{hamid2019smarttarget} by producing ground-truth deformations in $[5.58, 8.66]$ mm using the finite element modelling (FEM) process, proposed in
\cite{hu2011modelling,saeed2020prostate}, with different material properties assigned to peripheral zones (PZ) and transition zone (TZ): the ratios of Youngs' Modulus with PZ and TZ $\frac{E^{\text{PZ}}_k}{E^{\text{TZ}}_k}$ were in the range of $[0.10, 0.20]$. All data was resampled to isotropic resolutions being $1\times1\times1~\text{mm}^3$. For each patient, the pairs of point sets were randomly downsampled to $\mathbf{P}_\mathcal{S}$ and $\mathbf{P}_\mathcal{T}$ with  $N_s=N_t=1024$ and $N_s^{\text{surface}}=N_t^{\text{surface}}=512$ independently. 
In the first experiment, the root-mean-square error (\textsf{rmse}) of registration was defined between predicted displacement
and ground-truth $\mathbf{D}_\mathcal{S}^{gt}\in\mathbb{R}^{|\widetilde{N}_s|\times3}$ 
as
$   \textsf{rmse} =  \sqrt{ \frac{1}{|\widetilde{N}_s|}\sum_{s\in\widetilde{N}_s} ||\mathbf{d}_s-  \mathbf{d}_s^{\text{gt}}||_2^2  }$ where here $\widetilde{N}_s$ denote the set of all or surface points. In the second experiment, we computed the absolute percentage error (APE) as
$|(\text{ratio}_{\text{pred}}-\text{ratio}_{\text{gt}})/\text{ratio}_{\text{gt}}|$ for each case where $\text{ratio}_{\text{pred}}$ and $\text{ratio}_{\text{gt}}$ are predicted and ground-truth ratios, and also reported mean APE (mAPE) for all cases. Details about implementations such as network architectures are in the Supplementary material.\\ 
\noindent \textbf{Registration Performances} Fig. \ref{quantitative results of linear and nonlinear} includes the rmse values for all and surface points, respectively. Several key observations can be made from Fig. \ref{quantitative results of linear and nonlinear}: (1) except for the case with \textsf{Linear Cfg1} \cite{min2023non} for all points ($p=0.083$), all methods outperformed \textsf{Without PINNs} significantly at $\alpha=0.05$ level; (2) except for the case where \textsf{Linear Cfg1} and \textsf{Linear Cfg2} were significantly different for all points ($p=0.0496$), the differences between every pair of \textsf{PINNs} algorithms were not statistically significant; (3) \textsf{Nonlinear Cfg2} achieved the lowest mean registration error values being $1.25\pm0.62$ mm and $1.24\pm0.68$ mm for both all and surface points ($p<0.001$), against $1.80\pm0.44$ mm and $1.83\pm0.51$ mm \textsf{Without PINNs}. \\
\begin{figure}[t]
 \begin{tabular}{c}          
 \includegraphics[width=1\linewidth]{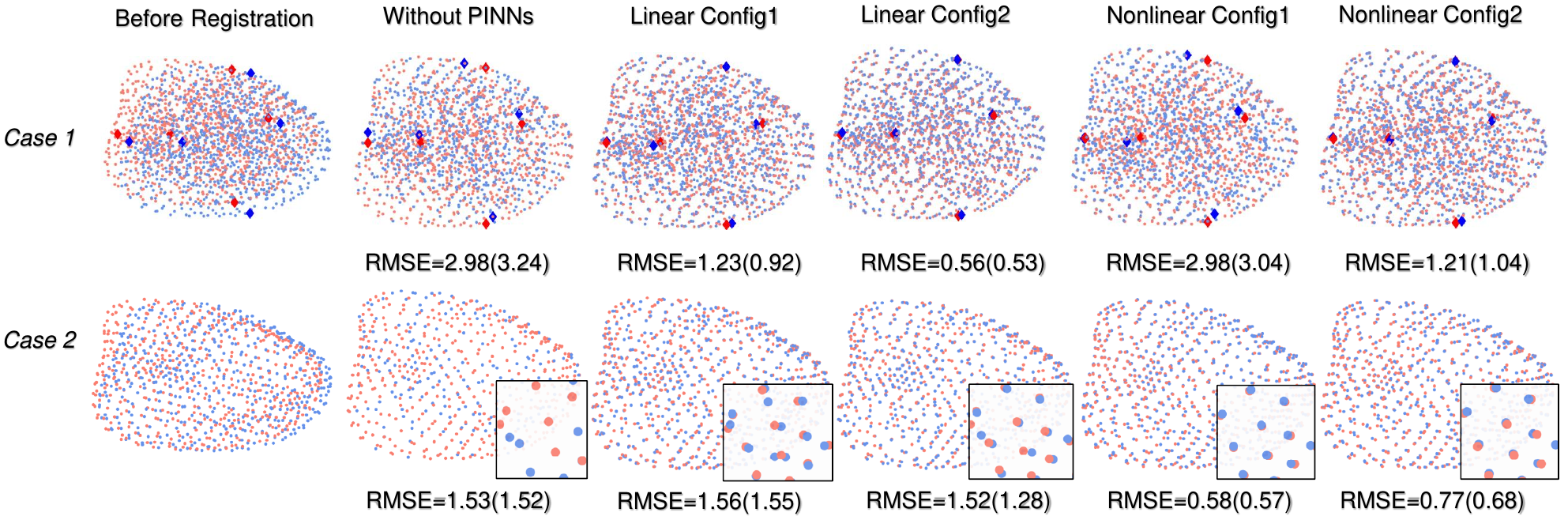}
\end{tabular}
\caption{Qualitative results for two cases with the five registration models. The rmse values for all points are also reported followed by those in parentheses for surface points.  
} 
\label{qualitative}
\end{figure}
\indent Fig. \ref{qualitative} shows the registration performances with two patient cases where blue and red dots respectively denote source (or predicted warped source) and target point sets with exact correspondences before (or after) registration. In \textit{Case 1}, the linear model (e.g., rmse values were 0.56 mm and 0.53 mm for all and surface points using \textsf{Linear Cfg2}) was better than the nonlinear model (e.g., rmse values
were 1.21 mm and 1.04 mm using \textsf{Nonlinear Cfg2}) and \textsf{Without PINNs} (i.e., rmse values were 2.98 mm and 3.24 mm), while nonlinear models outperformed linear ones and \textsf{Without PINNs} for \textit{Case 2}. In the first row of Fig. \ref{qualitative}, blue and red star shapes denote corresponding targets in two spaces. \\ 
\textbf{Results of the Inverse Problem} Fig. \ref{Inverse estimation optimisation process} demonstrates that the ratios of young's modulus between the PZ and TZ were recovered (only Cfg1s are reported due to much better performances than Cfg2s), by locating saddle points (i.e., through finding the flat line in Fig. \ref{Inverse estimation optimisation process}) during optimisation. 
The confidence interval is represented as shadow area in Fig. \ref{Inverse estimation optimisation process}. As shown in Fig. \ref{Inverse estimation optimisation process}, the nonlinear model exhibited significant advantages which is expected because more subtle (i.e., high-order) information  is preserved. Table \ref{Young's modulus ratio between the two regions} shows the ground-truth and predicted ratios for example cases, from which the APE was computed. The mAPE values with linear and nonlinear models were $76.28\%\pm30.97\%$ and $14.20\%\pm14.12\%$ respectively, indicating the nonlinear model performed better in the inverse problem ($p=0.002$). 
\begin{table}[t]
 \scriptsize
  \scriptsize
\centering
\caption{Young's modulus ratio between the two regions estimated with two models. 
}
\label{Young's modulus ratio between the two regions}
\begin{tabular}{p{40pt} p{50pt} p{60pt} p{50pt} p{60pt} p{50pt}  }\\
\hline
{\makecell[c]{Patient ID}}   & \makecell[c]{Ground-truth \\Ratio} &  {\makecell[c]{Predicted Ratio \\Linear}} &  \makecell[c]{APE\\Linear} & {\makecell[c]{Predicted Ratio\\ Nonlinear}} &  \makecell[c]{APE\\Nonlinear} \\
\hline
{\makecell[c]{Case 1}}&{\makecell[c]{0.19}}&{\makecell[c]{0.12}}&{\makecell[c]{36.84\%}}
&  {\makecell[c]{$\mathbf{0.19}$}}
& {\makecell[c]{$\mathbf{1.23\%}$}}\\ 
{\makecell[c]{Case 2}}&{\makecell[c]{0.16}}&{\makecell[c]{0.10}}&{\makecell[c]{37.50\%}}
&{\makecell[c]{$\mathbf{0.16}$}}
&{\makecell[c]{$\mathbf{1.25\%}$}} \\ 
{\makecell[c]{Case 3}}&{\makecell[c]{0.11}}&{\makecell[c]{0.22}}&{\makecell[c]{100.00\%}}
&
{\makecell[c]{$\mathbf{0.10}$}}
&{\makecell[c]{$\mathbf{9.09\%}$}} \\ 
{\makecell[c]{Case 4}}&{\makecell[c]{0.12}}&{\makecell[c]{0.22}}
& {\makecell[c]{83.33\%}}
&{\makecell[c]{$\mathbf{0.15}$}}
&{\makecell[c]{$\mathbf{25.00\%}$}} \\ 
{\makecell[c]{Case 5}}&{\makecell[c]{0.17}}&{\makecell[c]{0}}&{\makecell[c]{100.00\%}}
&{\makecell[c]{$\mathbf{0.19}$}}
&{\makecell[c]{$\mathbf{11.76\%}$}} \\ 
{\makecell[c]{Case 6}}&{\makecell[c]{0.19}}&{\makecell[c]{0}}
&{\makecell[c]{100.00\%}}
&
{\makecell[c]{$\mathbf{0.12}$}}
& {\makecell[c]{$\mathbf{36.84\%}$}} \\ \hline
\end{tabular}
\end{table}
\section{Discussions and Conclusions}
\label{discussions and conclusions}
We have demonstrated the success of incorporating the nonlinear elasticity in both \textsf{forward} and \textsf{inverse} problems of biomechanically constrained nonrigid point set registration. This work needs to be interpreted with limitations. First, the drawn conclusions may not be directly generalizable given the limited data size. Second, although results shed light on tackling the inverse problem along with the registration, it requires additional efforts (e.g., adding regularization terms) to further improve both accuracy and success rate. \\ 
\indent To conclude, in this paper, we have utilised PINNs to solve both the registration of soft tissues and the estimation of material properties, tackled as forward and inverse problems respectively, considering both linear and more complex nonlinear elasticities. Experimental results first show that no statistical significance is observed between linear and nonlinear models in the forward problem, among which results are highly variable across patients. The validity of adopting PINNs for solving estimating parameters, together with the superiority of the nonlinear model, is also demonstrated. These conclusions are drawn based on clinical data from prostate cancer patients, for topical applications including intraoperative motion modelling and multimodal image registration, potentially, new applications in better characterisation of pathological tissue motion.

\begin{credits}
\subsubsection{\ackname} This work was supported by the Wellcome/EPSRC Centre for Interventional and Surgical Sciences [203145Z/16/Z] and the International Alliance for Cancer Early Detection, an alliance between Cancer Research UK [C28070/A30912; C73666/A31378], Canary Center at Stanford University, the University of Cambridge, OHSU Knight Cancer Institute, University College London and the University of Manchester.  This work was also supported in part by the National Natural Science Foundation of China under Grant 62303275.
\subsubsection{\discintname}
The authors have no competing interests to declare that are
relevant to the content of this article.
\end{credits}
\bibliographystyle{splncs04}
\bibliography{Paper-0208}

\end{document}